\documentclass[letterpaper, 10 pt, conference]{ieeeconf}  

\IEEEoverridecommandlockouts                              
\newcommand{\argmin}{\mathop{\mathrm{argmin}}}

\newcommand{\TT}{\ensuremath{\mathsf{\scriptstyle{T}}}}
\newcommand{\T}{^{\TT}}

\newcommand{\inv}{^{\raisebox{.2ex}{$\scriptscriptstyle-1$}}}

\newcommand{\minus}{\scalebox{0.75}[1.0]{$-$}}

\newcommand{\idxk}{
\hspace{-0.5pt}(k)\hspace{0.5pt}
}

\newcommand{\idxkmo}{
\hspace{-0.5pt}(k\scalebox{0.75}[1.0]{\(-\)}1)\hspace{0.5pt}
}

\newcommand{\idxN}{
\hspace{-0.5pt}N\hspace{0.5pt}
}

\newcommand{\idxNmo}{
\hspace{-0.5pt}N\scalebox{0.75}[1.0]{\(-\)}1\hspace{0.5pt}
}

\usepackage{graphicx}
\usepackage{epstopdf}
\usepackage{epsfig} 
\usepackage{amsmath} 
\usepackage{amssymb}  
\usepackage{mathtools}
\usepackage{pgfplots}
\usepackage{amsfonts}
\usepackage{url}
\usepackage[]{units}
\usepackage{tikz}
\usepackage{multirow}
\usepackage{multicol}
\usetikzlibrary{calc}
\usetikzlibrary{positioning}

\pgfplotsset{compat=1.14}
\graphicspath{{./img/}}
\title{\LARGE \bf Iterative Learning Control for Fast and Accurate Position Tracking with an Articulated Soft Robotic Arm}

\author{Matthias Hofer, Lukas Spannagl and Raffaello D'Andrea
\thanks{The authors are members of the Institute for Dynamic Systems and Control, ETH Z\"urich, Switzerland. Email correspondence to Matthias Hofer {\tt\small hofermat@ethz.ch}.}}%
\begin{document}

\maketitle
\thispagestyle{empty}
\pagestyle{empty}
\begin{abstract}\label{sec:Abstract}
This paper presents the application of an iterative learning control scheme to improve the position tracking performance for an articulated soft robotic arm during aggressive maneuvers. Two antagonistically arranged, inflatable bellows actuate the robotic arm and provide high compliance while enabling fast actuation. Switching valves are used for pressure control of the soft actuators. A norm-optimal iterative learning control scheme based on a linear model of the system is presented and applied in parallel with a feedback controller. The learning scheme is experimentally evaluated on an aggressive trajectory involving set point shifts of 60 degrees within 0.2 seconds. The effectiveness of the learning approach is demonstrated by a reduction of the root-mean-square tracking error from 13 degrees to less than 2 degrees after applying the learning scheme for less than 30 iterations.
\end{abstract}

\section{Introduction}\label{sec:Introduction}
Soft, inflatable robotic manipulators exhibit a number of promising properties. High compliance and low inertia combined with pneumatic actuation enable fast, but nevertheless safe applications (\cite{BGorissen_ElasticInfl}, \cite{CBest_ControlOfAPneu}, \cite{SSanan_Physicalhum} and \cite{BMosadegh_PneumaticRapidly}).
However, accurate position control is challenging with soft manipulators, because they typically have a high number of potentially coupled and uncontrollable degrees of freedom. Furthermore, the dynamics of soft materials exhibit viscoelastic material behavior, which is difficult to model from first principles.

One way to improve the tracking performance of soft, inflatable manipulators is to combine soft structures with rigid components. These hybrid designs typically have lower overall compliance and higher inertia compared to their entirely soft counterparts, but also a reduced number of degrees of freedom \cite{RNatividad_AHybrid}. Furthermore, the degrees of freedom are limited to specific joints for articulated soft robots. Thereby, the control authority of the remaining degrees of freedom is generally higher, leading to an improved tracking performance. Examples for such designs can be found in \cite{HKim_Mechanical} and \cite{MJordan_PrecisePosition}.

Alternatively, advanced control approaches can be used to improve the tracking performance of inflatable robotic manipulators. Learning-based open loop control, which purely relies on mechanical feedback, is demonstrated in \cite{TThuruthel_LearningOpenloopControl} for a pneumatically-actuated soft manipulator. The authors of \cite{MGillespie_Simult} achieve a reduction in the tracking error by using model predictive control for a fabric-based soft arm. The control approach is extended in \cite{MGillespie_NonlinearMPC} to a nonlinear model predictive controller based on a neural network to describe the dynamics. Reinforcement learning is applied in \cite{YAnsari_AssistiveManipulator} to simultaneously optimize the accuracy of the controllable stiffness and the position tracking of a manipulator intended for assistive applications. A learned inverse kinematics model is used in \cite{RReinhart_InverseDynamicsFesto} to improve the position tracking accuracy with a soft handling assistant. The authors of \cite{FAngelini_ILCSoftInteraction} use iterative learning control (ILC) in combination with low-gain feedback control to improve tracking performance, while preserving the intrinsic compliance of soft robots. A method based on ILC to learn grabbing tasks for a soft fluidic elastomer manipulator is reported in \cite{AMarchese_DynTrajOpt}.
\begin{figure}
\centering
\includegraphics[trim=0mm 25mm 0mm 100mm, clip, width=\columnwidth]{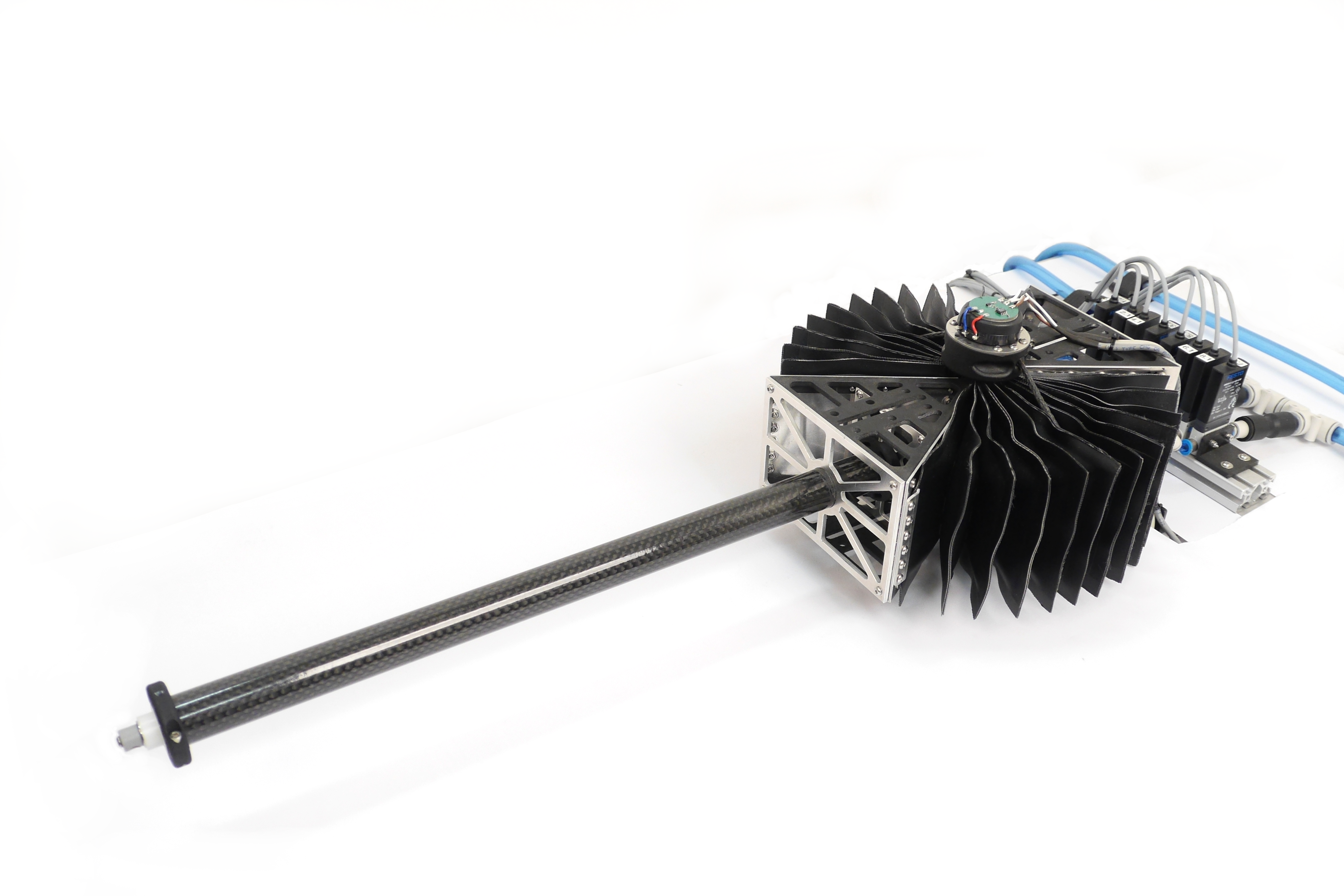}
\vspace{-18pt}
\caption{The articulated soft robotic arm used for the experimental evaluation. It consists of two antagonistically arranged soft bellow actuators and a rigid backbone. The inflatable actuators are made from coated fabric and attached to a lightweight structure built from a combination of 3D printed parts and aluminum plates. The low inertia of the one degree of freedom arm enables fast maneuvers.}
\vspace{-18pt}
\label{Fig:armPicture}
\end{figure}

In this paper, we propose an ILC scheme to improve the tracking performance for an articulated soft robotic arm (see Fig. \ref{Fig:armPicture}). Instead of the commonly used proportional valves (e.g. \cite{MJordan_PrecisePosition}, \cite{MGillespie_NonlinearMPC}), we deploy switching valves to control the pressure of the actuators (see also \cite{Hofer_DMC}). The hardware limitations imposed by this type of valve are addressed by a norm-optimal ILC approach, which is applied in parallel with a feedback controller. The non-causal learning controller is evaluated on repetitive, aggressive angle trajectories as arising in, for example, pick and place applications. 

The remainder of this paper is organized as follows: The design of the robotic arm is discussed in Section \ref{sec:SoftRoboticArm} along with the derivation of a simple model and a feedback controller. The iterative learning control approach is outlined in Section \ref{sec:LearningControl} and experimental results are presented in Section \ref{sec:Results}. Finally, a conclusion is drawn in Section \ref{sec:Conclusion}.

\section{Articulated Soft Robotic Arm}\label{sec:SoftRoboticArm}
The articulated soft robotic arm used for the experimental evaluation is presented in the first part of this section. It serves as a testbed for the study of control algorithms for this type of system. While the number of degrees of freedom is limited to one, it nevertheless exhibits the complex dynamics inherent in this type of system, e.g. the nonlinear pressure dynamics and viscoelastic material behavior. A simple model of the soft actuator and the robotic arm is derived in the second part and a cascaded control architecture is discussed in the last part of this section.
\subsection{Design}
The hybrid robotic arm consists of two soft bellow actuators and a rigid joint structure. The soft actuator is discussed in the first part and the complete robotic arm in the second part of this section. The design of the actuators and the robotic arm is inspired by the one presented in \cite{Hofer_DMC}. However, several properties of the actuators and the robotic arm are optimized.

\begin{figure}[h]
\centering
\includegraphics[trim=0mm 30mm 0mm 45mm, clip, width=\columnwidth]{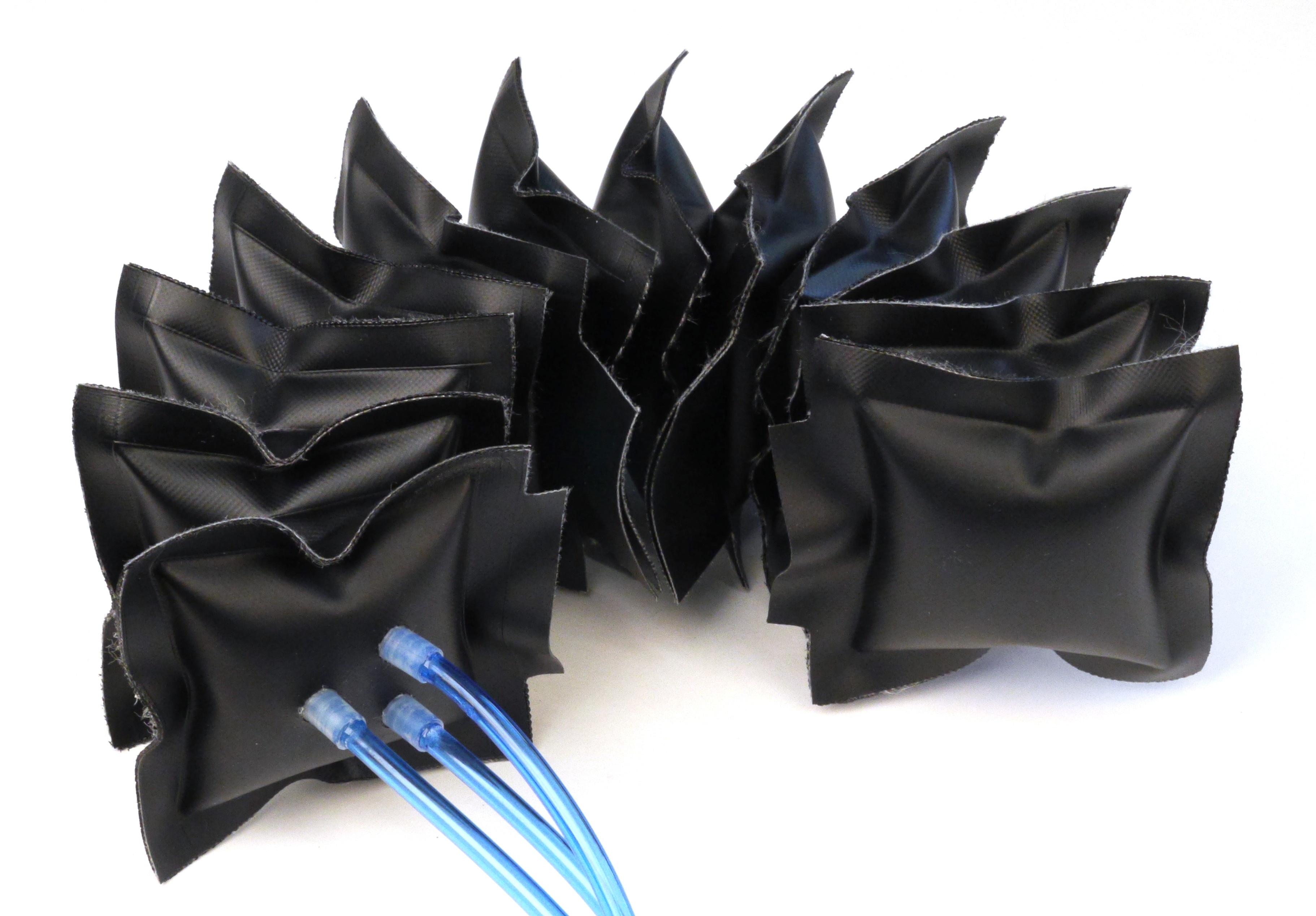}
\caption{The bellow-type actuator made from polyurethane coated nylon in fully inflated state. The actuator consists of twelve single cushions and the front side measures $\unit[100]\times\unit[100]{mm}$ in a deflated state. When fully inflated, the maximum actuation range, is approximately $\unit[190]{^\circ}$. Two tubes connect the actuator to the control valves, while the third one is used to measure the pressure in the actuator.}
\vspace{-6pt}
\label{Fig:actuator}
\end{figure}

A soft bellow actuator (Fig. \ref{Fig:actuator}) is made from thermoplastic polyurethane coated nylon (Rivertex\textregistered \ Riverseal\textregistered \ 842). It consists of twelve single cushions, which are connected by an internal seam. Placing the inner seam off-center leads to an expansion in the angular direction upon pressurization. The fabric pieces are prepared with a cutting plotter and processed with high frequency welding. Thereby, the fabric sheets are clamped between two electrodes and a high frequency alternating electromagnetic field is applied. The material coating is heated over its full seam thickness, eventually causing the fabric sheets to bond together (see \cite{HandbookOfPlasticJoining} for more details). Three high frequency weldable polyvinyl chloride tubing flanges are welded to each bellow. Tubing is glued to the flanges, which connect two tubes with the pressure regulating valves and the third one with the pressure sensor.
Instead of the commonly employed proportional solenoid valves, we use switching solenoid valves. While proportional valves can continuously adjust the nozzle, switching valves only have a binary state, namely fully open or fully closed. Continuous adjustment of the mass flow (and consequently of the pressure) is achieved by pulse width modulation (PWM). Four 2/2-way switching valves (Festo MHJ10) are used per actuator to control the pressure. Two valves are used as inlet valves, placed between the source pressure and the actuator, and two as outlet valves, placed between the actuator and the environment. The pressure is measured in each actuator and at the source using B\"urkert 8230 pressure transducers. The temperature is measured at the pressure supply using a temperature sensor (Texas Instruments LM35).

Note that the four deployed switching valves cost 2.5 times less than a proportional valve from the same manufacturer (Festo MPYE), which is commonly used in soft robotics applications (e.g. \cite{MJordan_PrecisePosition}, \cite{DBuechler_lightweightArm}). However, the advantage in price comes at the cost of a reduced flow capacity, which can be limiting for fast maneuvers. A control approach to address this limitation is presented in Section \ref{sec:LearningControl}.

With respect to the prototype presented in \cite{Hofer_DMC}, the actuation range is increased by using twelve instead of six single cushions. Furthermore, the pressure is directly measured in the actuator and not in the connecting tubes. Thereby, the actuator volume filters the effect of the pressure ripples caused by the switching valves and hence reduces valve jitter. Note that doubling the volume of the actuator requires the use of twice as many valves and connecting tubes in order to retain a comparable time constant of the pressure dynamics with respect to the smaller actuator.

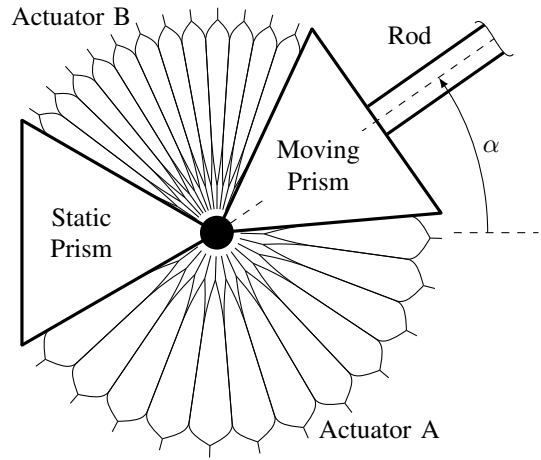
\begin{figure}[h]{}
   \centering
   \begin{tikzpicture}[scale = 0.45]

\def\alphaview{35}
\pgfmathsetmacro\angB{(120-\alphaview)/24}
\pgfmathsetmacro\angA{(120+\alphaview)/24}
\pgfmathsetmacro\startB{-(30+\angB)}
\pgfmathsetmacro\startA{(30+\angA)}
\pgfmathsetmacro\stepB{\startB-2*\angB}
\pgfmathsetmacro\stepA{\startA+2*\angA}
\pgfmathsetmacro\stopB{-(150-\alphaview)}
\pgfmathsetmacro\stopA{(150+\alphaview)}

\begin{scope}[shift = {(0,0)},scale = 0.7]
\foreach \angle in {\startB,\stepB,...,\stopB} {
\begin{scope}[rotate around = {\angle:(0,0)}]
\draw (-2,0) -- (-1,0);
\draw (-9.5,0) -- (-9,0);
\begin{scope}[rotate around = {-\angB:(0,0)}]
\draw (-8,0) -- (-4,0);
\end{scope}
\begin{scope}[rotate around = {\angB:(0,0)}]
\draw (-8,0) -- (-4,0);
\end{scope}
\draw  plot[smooth, tension=.9] coordinates {(-\angB:-4) ({-0.8*\angB}:-3) (-2,0)};
\draw  plot[smooth, tension=.9] coordinates {(\angB:-4) ({0.8*\angB}:-3) (-2,0)};
\draw  plot[smooth, tension=.9] coordinates {(-\angB:-8) ({-0.8*\angB}:-8.5) (-9,0)};
\draw  plot[smooth, tension=.9] coordinates {(\angB:-8) ({0.8*\angB}:-8.5) (-9,0)};
\end{scope}
}
\begin{scope}[rotate around = {-30:(0,0)}]
\draw[very thick] (0,0) -- (-9.5,0);
\node (A) at (-9.5,0){};
\end{scope}
\begin{scope}[rotate around = {30+\alphaview:(0,0)}]
\draw[very thick] (0,0) -- (9.5,0);
\node (B) at (9.5,0){};
\end{scope}
\end{scope}

\begin{scope}[shift = {(0,0)},scale = 0.7]
\foreach \angle in {\startA,\stepA,...,\stopA} {
\begin{scope}[rotate around = {\angle:(0,0)}]
\draw (-2,0) -- (-1,0);
\draw (-9.5,0) -- (-9,0);
\begin{scope}[rotate around = {-\angA:(0,0)}]
\draw (-8,0) -- (-4,0);
\end{scope}
\begin{scope}[rotate around = {\angA:(0,0)}]
\draw (-8,0) -- (-4,0);
\end{scope}
\draw  plot[smooth, tension=.9] coordinates {(-\angA:-4) ({-0.8*\angA}:-3) (-2,0)};
\draw  plot[smooth, tension=.9] coordinates {(\angA:-4) ({0.8*\angA}:-3) (-2,0)};
\draw  plot[smooth, tension=.9] coordinates {(-\angA:-8) ({-0.8*\angA}:-8.5) (-9,0)};
\draw  plot[smooth, tension=.9] coordinates {(\angA:-8) ({0.8*\angA}:-8.5) (-9,0)};
\end{scope}
}
\begin{scope}[rotate around = {30:(0,0)}]
\draw[very thick] (0,0) -- (-9.5,0);
\node (C) at (-9.5,0){};
\end{scope}
\begin{scope}[rotate around = {-30+\alphaview:(0,0)}]
\draw[very thick] (0,0) -- (9.5,0);
\node (D) at (9.5,0){};
\end{scope}
\end{scope}

\draw[very thick] (A.center) -- (C.center);
\draw[very thick] (B.center) -- (D.center);
\fill[black] (A.center) circle (0.042);
\fill[black] (B.center) circle (0.042);
\fill[black] (C.center) circle (0.042);
\fill[black] (D.center) circle (0.042);
\fill[black] (0,0) circle (0.5);

\begin{scope}[rotate around = {\alphaview:(0,0)}]
\node (RodAnc) at (10,-0.5){};
\draw[dashed] (0,0) -- (10,0);
\draw[very thick] (5.8,0.5) -- (10, 0.5);
\draw[very thick] (5.8,-0.5) -- (10, -0.5);
\draw  plot[smooth, tension=.8] coordinates {(9.8,0.7)(10, 0.5)  (10, -0.5) (10.2,-0.7)};
\end{scope}

\draw[dashed] (7,0) -- (9.5,0);
\draw[-latex] (0,0)++(0:8) arc (0:\alphaview:8);
\begin{scope}[rotate around = {\alphaview/2:(0,0)}]
\node at (8.5,0){$\alpha$};
\end{scope}
\node (AcB) at (-4.3, 6.4) {Actuator B};
\node (AcA) at (4.8,-5.8) {Actuator A};
\node[fill=white, align = center] (SPr) at (-4.0,0.0) {Static \\ Prism};
\node[fill=white, align = center] (MPr) at (3.0,2.0) {Moving \\ Prism};
\node[fill=white, align = center] (Rod) at (5.7,5.85) {Rod};
\end{tikzpicture}
   \caption{Schematic drawing of the articulated soft robotic arm including two actuators, namely A and B, the static and moving prisms, the rod and the revolute joint indicated by the black circle. In the configuration shown, the pressure in actuator A is higher than in actuator B, leading to a deflection in the positive $\alpha$-direction.}
  \label{fig:SoftRoboticArmSchmematic}
\end{figure}

The hybrid robotic arm consists of two bellow-type actuators, which are arranged antagonistically between a rigid support structure consisting of two prisms (see Fig. \ref{fig:SoftRoboticArmSchmematic}). A prism consists of two 3D printed triangles (polyamide PA12), which are connected by aluminum plates. The first and last cushion of each actuator are attached to the aluminum plates of either the static or the moving prism. The two prisms are connected by revolute joints, which are integrated into the triangles. The angle of the arm is measured by a rotary encoder (Novotechnik P4500) providing angle data with an accuracy to one-tenth of a degree. Finally, a carbon fiber rod is attached to the moving prism to form the arm.  

A pressure difference between actuators A and B leads to a different expansion of the two actuators and consequently to a torque on the moving prism. In order to maximize the expansion of each actuator in the angular direction and hence minimize the expansion in the radial direction, a strap is attached around both actuators.

As a consequence of the increased angle range compared to the design presented in \cite{Hofer_DMC}, the spring-like retraction forces of the deflated actuators are reduced. Therefore, the required actuation pressure for a given angle is reduced as well. The combination of 3D printed plastic parts with aluminum plates considerably reduces the inertia of the robotic arm compared to the previous design. The hybrid design presented combines the benefits of soft actuators, namely high compliance and fast actuation, with the advantages of classic robotic arms, namely that the expansion of the soft actuators is concentrated in a single degree of freedom.
\subsection{Modeling}
The modeling of the robotic arm is described in this section. It is important to note that, for all models discussed here, the focus lies on simplicity rather than accuracy. The motivation is to capture the principle dynamics and compensate for inaccuracies and model uncertainties by a learning-based control approach. 

The model of the robotic arm can be divided into two subsystems. The pneumatic system including the soft actuator and the valves is described first and the rigid body dynamics of the robotic arm are discussed in a second part. 

The pressure dynamics of the actuator can be derived from first principles. Taking the time derivative of the ideal gas law relates the change in pressure to the mass flow in or out of the actuator and the change in volume and temperature. We assume isothermal conditions and neglect changes in the volume, which leads to the following expression for the pressure dynamics of an actuator,
\begin{equation}
\dot{p} = \dot{m}\frac{RT}{V} +\dot{T}\frac{mR}{V}-\dot{V}\frac{p}{V} \approx \dot{m}\frac{RT}{V}, \label{eq:pressDyn}
\end{equation}
where $R$ denotes the ideal gas constant, $m$ the air mass in the actuator, $\dot{m}$ the mass flow in or out of the actuator, $T$, $\dot{T}$ the temperature and its derivative and $V$ the volume and $\dot{V}$ its derivative. Given the isochoric assumption, the volume of the actuator is approximated as constant, hence independent of the angle of the robotic arm. The average volume is measured to be $\unit[0.45]{L}$. Note that the model presented here neglects interactions with the antagonistic actuator and the elasticity of the fabric material.

More sophisticated models of the pressure dynamics including an identification of the angle dependent volume relation were also investigated. However, it only slightly improved the pressure control performance, but required additional identification experiments. Therefore, we only use a coarse approximation of the pressure dynamics in this work and rely on the learning approach to compensate for the approximation.

A static model of the valves employed is derived in \cite{Hofer_DMC}, which is also used in this work. The experimentally identified model relates the mass flow of one actuator to the up and downstream pressures of its valves and to the applied duty cycles,
\begin{equation}
\dot{m} =   f_{\text{valves}}(p_{\text{u}}, p_{\text{d}}, dc_{\text{in}}, dc_{\text{out}}),
            \label{eq:valveMdl}
\end{equation}
where $p_{\text{u}}$ and $p_{\text{d}}$ denote up and downstream pressure and $dc_{\text{in}}$, $dc_{\text{out}}$ the duty cycles of the inlet and outlet valves. For the inlet valves, the upstream pressure corresponds to source pressure and the downstream pressure to the controlled pressure in the actuator. For the outlet valves, upstream pressure is related to the pressure in the actuator and downstream pressure to ambient pressure.

The rigid body dynamics of the robotic arm are assumed to be driven by the pressure difference between the two actuators, defined as $\Delta p=p_{\text{A}}-p_{\text{B}}$. A positive pressure difference, $\Delta p$, accelerates the arm in the positive $\alpha$-direction (see Fig. \ref{fig:SoftRoboticArmSchmematic}). The dynamics of the robotic arm with $\Delta p$ as input and the arm angle $\alpha$ as output, are identified using system identification. The same identification procedure as in \cite{Hofer_DMC} is applied. A continuous-time second-order model is assumed for the dynamics of the robotic arm. More complex models including the actuator pressures as states and the set point pressure difference as an input are not considered for the sake of simplicity and will be compensated by the learning approach. The resulting transfer function is,
\begin{equation}
G(s) = \frac{\alpha (s)}{\Delta p(s)}= \kappa\frac{\omega^2_0}{\omega^2_0+2\delta \omega_0 s+s^2}.\label{eq:TF}
\end{equation}
The complex variable is denoted by $s$ and the numeric values of the parameters are $\kappa=\unit[7.91]{rad/bar}$, $\omega_0=\unit[14.14]{1/s}$ and $\delta=\unit[0.31]{}$. The design improvements discussed in the previous subsection are reflected in the parameter values identified. Compared to the model presented in \cite{Hofer_DMC}, the gain of the transfer function, $\kappa$, is more than four times larger due to the reduced spring-like retraction forces. The continuous time model is discretized with a sampling time of $\unit[1/50]{s}$ leading to the following linear-time-invariant system,
\begin{align}
\begin{bmatrix}
       x_1\idxk\\[0.3em]
       x_2\idxk
\end{bmatrix}
&=
\underbrace{
\begin{bmatrix}
       0.96 & 0.18\\[0.3em]
       -0.36 & 0.80\\
\end{bmatrix}
}_{\coloneqq A}
\begin{bmatrix}
       x_1\idxkmo\\[0.3em]
       x_2\idxkmo\\
\end{bmatrix}
+
\underbrace{
\begin{bmatrix}
       0.09\\[0.3em]
       0.91
\end{bmatrix}
}_{\coloneqq B}
u\idxkmo \nonumber\\
y\idxk
&=
\underbrace{
\begin{bmatrix}
       1 & 0\\
\end{bmatrix}
}_{\coloneqq C}
\begin{bmatrix}
       x_1\idxk\\[0.3em]
       x_2\idxk
\end{bmatrix},
\label{eq:LTIproc}
\end{align}
where $k$ denotes the time index, $(x_1, x_2)$ corresponds to the state $(\alpha, \dot{\alpha})$ normalized by $(\pi, \unit[10\pi]{1/s})$ and $u$ to the control input $\Delta p$ (in Pa) normalized by $\unit[1\mathrm{e}{5}]{Pa}$. The arm deflection $\alpha$ is directly measured by the rotary encoder. This model will be used for the iterative learning control discussed in Sec. \ref{sec:LearningControl}.


\subsection{Control}
A cascaded control architecture similar to \cite{Hofer_DMC} is applied based on time scale separation of the faster pressure dynamics in an inner loop and the slower arm dynamics in an outer loop (see Fig. \ref{fig:blockdiagram}). A proportional-integral-derivative (PID) controller is used in the outer loop to compute a required pressure difference, $u_{\text{PID}}$, to reach a desired arm angle. The PID controller includes a feed forward term to improve tracking performance and has the following form,
\begin{equation}
u_{\text{PID}} = k_{\text{ff}} y_{\text{D}} + k_{\text{p}}(y_{\text{D}}-y^j) + k_{\text{i}}\int (y_{\text{D}}-y^j) dt + k_{\text{d}} \dot{y}^j,
\label{eq:PID}
\end{equation}
\vspace{-2pt}
where the time index is omitted for the ease of notation. The desired angle is denoted by $y_{\text{D}}$ and the measured angle by $y^j$. The set point for the pressure difference is then translated to the individual pressure set points for the inner control loops. The underlying assumption is that set point changes of the pressure can be tracked instantaneously by the inner control loop. From the control input determined by \eqref{eq:PID}, the set point pressures are computed as,
\vspace{-2pt}
\begin{align}
\begin{split}
p_{\text{A,des}} &= \min(p_{\text{max}}, \max(p_{\text{0}},p_{\text{0}}+u_{\text{PID}}))\\
p_{\text{B,des}} &= \min(p_{\text{max}}, \max(p_{\text{0}},p_{\text{0}}-u_{\text{PID}})),\label{eq:PID2Setpoints}
\end{split}
\end{align}
\vspace{-2pt}
where $p_{\text{0}}$ is the ambient air pressure and $p_{\text{max}}$ the maximum allowed pressure.

Next, we present the pressure controller based on the model presented in the previous subsection. We describe the control algorithm for actuator A, but it is analogously implemented for actuator B. In order to smooth the effects of the PWM, a moving average filter is used to filter the measured pressures and hence reduce valve jitter. We assume a first order model for the pressure dynamics,
\vspace{-2pt}
\begin{equation}
\dot{p}_{\text{A}}=\frac{1}{\tau_{\text{p}}}(p_{\text{A,des}}-p_{\text{A}}), \label{eq:desPressDyn}
\end{equation}
\vspace{-2pt}
where $\tau_{\text{p}}$ is the time constant of the desired pressure dynamics and is used as a tuning parameter for the controller. We combine this expression with \eqref{eq:pressDyn} to obtain the desired mass flow $\dot{m}$ as a function of the current pressure deviation,
\vspace{-4pt}
\begin{equation}
\dot{m}_{\text{A}} =  \frac{V}{RT\tau_{\text{p}}}(p_{\text{A,des}}-p_{\text{A}}).
      \label{eq:massFlowControl}
\end{equation}
\vspace{-2pt}
The valve model \eqref{eq:valveMdl} is used to compute the required duty cycles for both the in and outlet valves from the desired mass flow and the known up and downstream pressures of the valves. A detailed explanation of this procedure can be found in \cite{Hofer_DMC}. Note that the number of valves is doubled compared to \cite{Hofer_DMC}, but the same duty cycles are applied to both in and both outlet valves of actuator A and B.

\section{Iterative Learning Control}\label{sec:LearningControl}
In this section, we present an ILC scheme to improve the angle tracking performance with the robotic arm. ILC is a method to improve the tracking performance for repetitive tasks (\cite{DBristow_ASurveyOfIter}). The tracking error from the previous iteration is used to compensate for the disturbances in the current iteration.

\begin{figure}[]
\tikzset{block/.style={draw,inner sep = 3pt,minimum height=.7cm,minimum width=1cm},
         adder/.style={draw,circle,inner sep=2pt},
         dot/.style={draw,circle,inner sep=0.05pt,fill=black},
         >=stealth}

\begin{center}
\begin{tikzpicture}[xscale = 1.0, yscale=1.0]
\node[] (left_a1) at (0,0) {};
\node[adder,label=-100:$\minus$, right of=left_a1, node distance=0.7cm](a1) {};
\node[dot,label={[xshift=-2mm, yshift=3mm]$e^{j}$}, right of=a1, node distance=0.4cm] (right_a1) {};
\node[block, right=0.2cm of right_a1] (controller) {$\ensuremath{\textup{PID}}$};
\node[adder,label={[xshift=-2mm, yshift=2mm]$u^{j}$}, right of=controller, node distance=1.0cm](a2) {};
\node[dot, right of=a2, node distance=0.3cm] (right_a2) {};
\node[block, above right=0.1cm and 0.5cm of a2, node distance=1.5cm] (pCtrlA) {$\ensuremath{\textup{pCtrl A}}$};
\node[block, below right=0.1cm and 0.5cm of a2, node distance=1.5cm] (pCtrlB) {$\ensuremath{\textup{pCtrl B}}$};
\node[block, right=0.3cm of pCtrlA] (ActuatorA) {$\ensuremath{\textup{Act A}}$};
\node[block, right=0.3cm of pCtrlB] (ActuatorB) {$\ensuremath{\textup{Act B}}$};
\node[block, right of=a2, node distance=4.4cm] (plant) {$\ensuremath{\textup{Arm}}$};
\node[right of=plant, node distance=1.3cm] (right_plant) {};
\node[block, above right=0.8cm and 0.2cm of right_a1] (ILC) {$\ensuremath{\textup{NOILC}}$};
\draw[-] (a2) -- (right_a2);
\draw[->] (right_a2) |- ([shift={(0mm,-2mm)}]pCtrlA);
\draw[->] (right_a2) |- ([shift={(0mm,2mm)}]pCtrlB);
\draw[->] (pCtrlA) -- (ActuatorA);
\draw[->] (pCtrlB) |- (ActuatorB);
\node[dot, right of=ActuatorA, node distance=0.854cm] {};
\node[dot, right of=ActuatorB, node distance=0.885cm] {};
\draw[->] (ActuatorA) -|([shift={(3mm,-0.5mm)}]ActuatorA.south east)-- ([shift={(0mm,-2.3mm)}]plant.north west);
\draw[->] (ActuatorB) -|([shift={(3.4mm,0.5mm)}]ActuatorB.north east)-- ([shift={(0mm,2.3mm)}]plant.south west);
\draw[->] ([shift={(3mm,0mm)}]ActuatorA.east)-- ([shift={(3mm,10mm)}]ActuatorA.south east)-- ([shift={(-30.1mm,10mm)}]ActuatorA.south east) |-([shift={(0mm,-2mm)}]pCtrlA.north west);
\draw[->] ([shift={(3.4mm,0mm)}]ActuatorB.east)-- ([shift={(3.4mm,-10mm)}]ActuatorB.north east)-- ([shift={(-29.7mm,-10mm)}]ActuatorB.north east) |-([shift={(0mm,2mm)}]pCtrlB.south west);
\draw[->] (left_a1) -- (a1) node[midway, above] {$y_{\text{D}}$};
\draw[->] (a1) -- (controller) node[midway, below] {};
\draw[->] (right_a1) |- (ILC); 
\draw[->] (controller) -- (a2); 
\node[dot, right of=plant, node distance=8.08mm] {};
\draw[->] ([shift={(3mm,0mm)}]plant.east) |- ([shift={(0mm,-13mm)}]a1.south)  -- (a1);  
\draw[->] (plant) -- (right_plant) node[midway, above] {$y^j$}; 
\draw[->] (ILC) -| (a2); 
\node[above right=-1mm and 2.5mm of ActuatorA] (pA) {$p_{\text{A}}$};
\node[below right=-0.8mm and 2.8mm of ActuatorB] (pB) {$p_{\text{B}}$};
\node[below right=-3.5mm and -0.8mm of controller] (uPID) {$u_{\text{PID}}$};
\end{tikzpicture} 
\end{center}
  \vspace{-6pt}
  \caption{Control architecture with the NOILC scheme in parallel configuration with the PID feedback controller in the outer control loop. The feed forward signal of the PID controller is not depicted for the sake of clarity. The input to the NOILC is the normalized error between the desired, $y_{\text{D}}$, and measured angle, $y^j$. The current value of the correction signal $u^j$ is added to the control input of the feedback controller, $u_{\text{PID}}$, and forms the input to the inner control loops. The pressure controllers (pCtrl A, pCtrl B) adjust the pressures in each actuator (Act A, Act B). These pressures are the inputs to the system (Arm).}
  \vspace{-16pt}
  \label{fig:blockdiagram}
\end{figure}
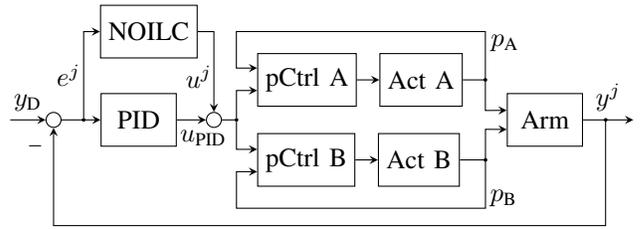

We apply a norm-optimal iterative learning control (NOILC) scheme (see \cite{Owens_ILC}), which is based on the linear model derived in Section \ref{sec:SoftRoboticArm} and a quadratic cost function. Since the NOILC approach can only account for repetitive disturbances, it is applied in parallel with the previously presented feedback controller, which compensates for non-repetitive disturbances (as discussed in \cite{DBristow_ASurveyOfIter}). The resulting control architecture is depicted in Fig. \ref{fig:blockdiagram}, where the superscript index $j$ is used to denote an iteration of the learning scheme. The tracking error in iteration $j$ is used to compute a correction signal, which is added to the feedback controller in the next iteration. 
The lifted-system framework (see \cite{DBristow_ASurveyOfIter}) is used to derive the NOILC algorithm. The following variables are introduced,
\vspace{-4pt}
\begin{align}
y_{\text{D}}   &:= \begin{pmatrix}y_{\text{D}}(1), \dots, y_{\text{D}}(\idxN)\end{pmatrix} & d &:= \begin{pmatrix}d(1), \dots, d(\idxN)\end{pmatrix} \nonumber\\
y^j &:= \begin{pmatrix}y^j(1), \dots, y^j(\idxN)\end{pmatrix} & u^j &:= \begin{pmatrix}u^j(0), \dots, u^j(\idxNmo)\end{pmatrix} \nonumber\\
e^j   &:= \begin{pmatrix}e^j(1), \dots, e^j(\idxN)\end{pmatrix} && \text{all} \in \mathbb{R}^N, \label{eq:liftedsystem}
\end{align}
where $y_{\text{D}}$ is the desired output and $y^j$ the measured output in iteration $j$. The normalized error in iteration $j$ is denoted by $e^j$, the repetitive disturbance by $d$ and the correction signal applied in iteration $j$ by $u^j$. The desired output $y_{\text{D}}$ is the same for all iterations and the disturbance $d$ is assumed to be repetitive between successive iterations. Note that four variables in \eqref{eq:liftedsystem} are shifted by one time step to account for the one step time delay induced by the plant. The dimension of the lifted-system, that is the number of time steps, $N$, is defined by the time duration of the desired trajectory and the sampling time of the NOILC approach, $T_{\text{ILC}}$.

Defining the lifted state matrix, 
\vspace{-4pt}
\begin{equation}
P = 
\setlength\arraycolsep{2pt}
\begin{bmatrix}
       CB  & 0 & 0 & 0\\[0.3em]
       \vdots & & CB & 0\\[0.3em]
       CA^{\idxNmo} B & \cdots & CAB & CB\\
\end{bmatrix}\in \mathbb{R}^{N\times N},\label{eq:P}
\end{equation}
allows us to express the dynamics in the lifted system framework, namely,
\vspace{-4pt}
\begin{equation}
\begin{aligned}
y^j &= P u^j + d\\
e^j &= y_{\text{D}} - y^j = y_{\text{D}} - P u^j - d\\
e^{j+1} &= y_{\text{D}} - P u^{j+1} - d\\
        &= e^{j}-P (u^{j+1}-u^{j}),\label{eq:dynamcisliftedsystem}
\end{aligned}
\end{equation}
where we assume a zero initial condition for the state. The principle idea in NOILC is to obtain the correction input by minimizing a quadratic cost function of the predicted tracking error of the next iteration. The linear model in \eqref{eq:dynamcisliftedsystem} is used to predict the tracking error one iteration ahead. Additional terms can be added to the cost function to improve the transient learning behavior. The following objective function is used in this work, 
\vspace{-2pt}
\setlength{\belowdisplayskip}{2pt} \setlength{\belowdisplayshortskip}{0pt}
\setlength{\abovedisplayskip}{2pt} \setlength{\abovedisplayshortskip}{0pt}
\begin{align}
J(u^{j+1}) = \frac{1}{2} [ &e^{{j+1}^\TT} M e^{j+1} + (u^{j+1}\minus u^{j})\T S (u^{j+1}\minus u^{j}) \nonumber\\
&+ u^{{j+1}^\TT} D\T WD u^{j+1}],
\label{eq:objective}
\end{align}
where $M$, $S$, $W$ are positive semi-definite cost matrices of appropriate dimensions and 
\vspace{-2pt}
\begin{equation}
D = \frac{1}{T_{\text{ILC}}}
\setlength\arraycolsep{2pt}
\begin{bmatrix}
       \minus1 & 1 & & \multicolumn{2}{c}{\multirow{2}{*}{\Huge$0$}} \\[0.3em]
        &  \ddots & \ddots  &  & \\[0.3em]
        \multicolumn{2}{l}{\multirow{2}{*}{\Huge$0$}} & & \mkern-40mu \minus1 & \mkern-10mu 1\\[0.3em]
        &  &  & \mkern-40mu \minus1 &  \mkern-10mu 1\\
\end{bmatrix}\in \mathbb{R}^{N\times N}\label{eq:D}
\vspace{-2pt}
\end{equation}
approximates the derivative of the correction input, by the first order Euler forward numerical differentiation. The first term in \eqref{eq:objective} penalizes the predicted tracking error in the next iteration and can be replaced by the last expression in \eqref{eq:dynamcisliftedsystem}. The second term penalizes a change in the correction input between successive iterations and the last term the derivative of the correction input. These last two terms are included to ensure a stable learning behavior by avoiding an overcompensation of the disturbance. Including a cost on the absolute change in correction input ensures that the NOILC approach does not excessively compensate for the previous tracking error before the computed correction input can show effect. Penalizing the derivative of the correction input limits the high frequency content of the correction input, which is handled in traditional ILC approaches by the so-called Q filters (see \cite{DBristow_ASurveyOfIter}). Constraints on the correction input are neglected in the minimization of \eqref{eq:objective}, such that the optimal solution can be computed in closed form. The feasibility of the solution is imposed by limiting the resulting pressure set points in \eqref{eq:PID2Setpoints} to the feasible interval $[p_{\text{0}}, p_{\text{max}}]$ at a later stage. The optimal solution is therefore given by,
\setlength{\belowdisplayskip}{5pt} \setlength{\belowdisplayshortskip}{0pt}
\setlength{\abovedisplayskip}{5pt} \setlength{\abovedisplayshortskip}{0pt}
\begin{align}
u^{j+1^{\star}} &= \argmin_{u^{j+1}} J(u^{j+1})\nonumber\\
&= (P\T MP+S+D \T WD)\inv(P \T MP+S) u^{j}\nonumber\\
&\quad+ (P\T MP+S+D \T WD)\inv P\T M e^{j}.
\label{eq:ustar}
\end{align}
Note that the inverse in \eqref{eq:ustar} exists if the cost matrix $S$ has full rank. The correction input of the NOILC approach is added to $u_{\text{PID}}$ as given in \eqref{eq:PID} for each time step,
\begin{equation}
u_{\text{tot}}^{j}(k) = u_{\text{PID}}(k)+u^{j}(k), \quad k = 1,\dots,N, 
\end{equation}
and consequently $u_{\text{tot}}^j$ is used in \eqref{eq:PID2Setpoints} instead of $u_{\text{PID}}$. 

Simpler ILC approaches (e.g. model-free), such as PD-type ILC (see \cite{DBristow_ASurveyOfIter}), were also investigated. However, contrary to the discussed NOILC approach they converged for very aggressive trajectories (compare Sec. \ref{sec:Results}) only if the high-frequency content of the correction signal was suppressed significantly by tuning a Q filter accordingly. However, the Q filter clearly limited the possible improvement of the tracking performance.

The previously introduced approach requires to store a correction signal and we finish this section by investigating how much that costs. Assuming a price of $0.03$ USD per gigabyte of permanent hard disk drive memory as of 2018 \cite{memPrice}, a precision of two bytes to store the final correction signal, and a sampling time $T_{\text{ILC}}$ as stated in the next section, it costs $1$ USD to store $10.5$ years of correction signal.

\vspace{-14pt}
\section{Experimental Results}\label{sec:Results}
\vspace{-3pt}
In this section, we present the results from experimental evaluations of the proposed NOILC scheme on the articulated soft robotic arm.

The control algorithms are executed on a laptop computer (Intel Core i7 CPU, \unit[2.8]{GHz}) and the valves and sensors are interfaced over a Labjack T7 Pro device. The pressure controllers are executed at $\unit[200]{Hz}$, the PID feedback controller runs at $\unit[50]{Hz}$ and the NOILC scheme has a sampling time of $T_{\text{ILC}} = \unit[1/50]{s}$. The source pressure is set to $\unit[3]{bar}$, the maximum pressure constraint to $p_{\text{max}}=\unit[1.4]{bar}$, the time constant of the pressure controllers to $\tau_{\text{p}}=\unit[1/50]{s}$ and the PWM frequency is set to $\unit[200]{Hz}$. The cost matrices of the NOILC scheme are $M = I_N$, $S = 0.1 \cdot I_N$ and $W = 2\mathrm{e}{-5}\cdot I_N$ with $I_N\in \mathbb{R}^{N\times N}$ being the identity matrix. 

The reference trajectory to evaluate the learning scheme has a duration of $\unit[8]{s}$ (resulting in $N=400$) and includes several set point jumps of $\unit[60]{^\circ}$ within $\unit[0.2]{s}$ and peak angular accelerations of $\unit[12000]{^\circ/s^2}$. It is computed from trapezoidal angular velocity profiles, which limit the maximum angular acceleration and correspondingly the excited frequency band. The reader is referred to the video attachment to gain an impression of the experiments conducted.

\begin{figure}[]
\input{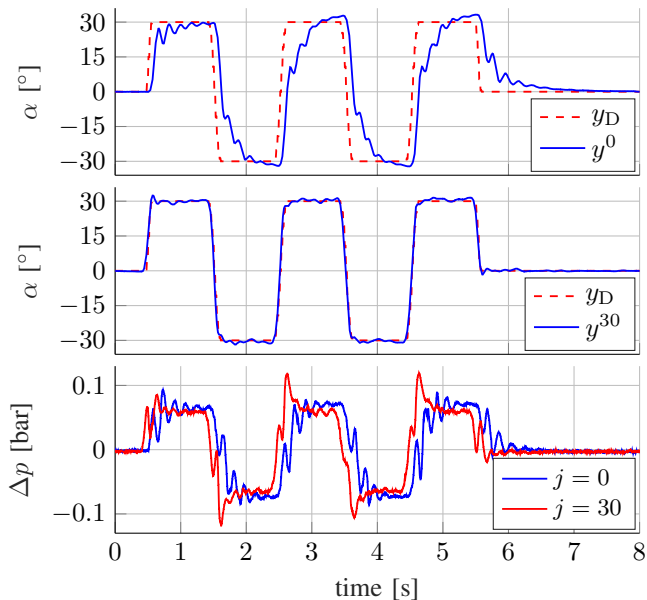}
\vspace{-12pt}
\caption{Experimental results of the robotic arm tracking an angular set point trajectory. The top plot shows the tracking performance when the feedback controller is used only (no learning, iteration 0). In this case, the angle initially follows the set point with a steep response, then slightly bounces back and finally reaches the set point. The middle plot shows the improved tracking performance after applying the NOILC scheme for 30 iterations. The bottom plot depicts the total pressure difference applied in iteration 0 (blue) and iteration 30 (red) and reveals the importance of the non-casual nature of the NOILC scheme, resulting in a shifted pressure difference input anticipating the repetitive disturbances. For the entire trajectory, the total pressure difference is never close to its maximum pressure constraint.}
\label{fig:iterComp}
\vspace{-6pt}
\end{figure}

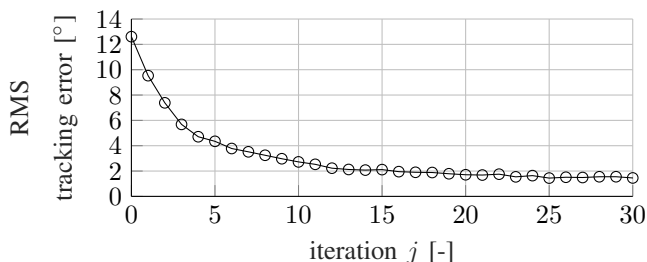
\begin{figure}[]
%
%
\begin{tikzpicture}

\begin{axis}[%
width=2.624in,
height=0.929in,
at={(0.581in,0.42in)},
scale only axis,
xmin=0,
xmax=30,
xlabel style={font=\color{white!15!black}},
xlabel={iteration $j$ [-]},
ymin=0,
ymax=14,
ytick={ 0,  2,  4,  6,  8, 10, 12, 14},
ylabel style={font=\color{white!15!black}, align=center},
ylabel={RMS\\[1ex]tracking error [$^{\circ}$]},
axis background/.style={fill=white},
axis x line*=bottom,
axis y line*=left,
xmajorgrids,
ymajorgrids
]
\addplot [color=black, mark=o, mark options={solid, black}, forget plot]
  table[row sep=crcr]{%
0	12.6077598446695\\
1	9.52917869070464\\
2	7.38511801773115\\
3	5.68307029519881\\
4	4.70852418425094\\
5	4.3467921377098\\
6	3.77909588191771\\
7	3.51853175604186\\
8	3.24621970865172\\
9	2.97229198798794\\
10	2.72266023752891\\
11	2.53074580935831\\
12	2.22603232595981\\
13	2.12434592113773\\
14	2.0753545639041\\
15	2.11344520587076\\
16	1.95283610875782\\
17	1.90213577465065\\
18	1.88433807368623\\
19	1.78711665865174\\
20	1.70818691219536\\
21	1.68514978028307\\
22	1.75288449328185\\
23	1.542640290293\\
24	1.6387494971834\\
25	1.44489043635098\\
26	1.51094666219736\\
27	1.49016413529381\\
28	1.54959976652334\\
29	1.55140162459567\\
30	1.45516338926097\\
31	1.47908985098068\\
};
\end{axis}
\end{tikzpicture}%
\vspace{-12pt}
\caption{The RMS tracking error plotted over 30 iterations of applying the NOILC scheme. The error decreases from $\unit[13]{^\circ}$ to less than $\unit[2]{^\circ}$ following a monotonic trend and remains constant thereafter. I.e. no early stopping is required. Deploying the learning approach for only 10 iterations already accounts for $\unit[90]{\%}$ of the improvement achieved after 30 iterations.}
\label{fig:error}
\vspace{-18pt}
\end{figure}

The results of the angle tracking experiment are shown in Fig. \ref{fig:iterComp}. The tracking performance is improved significantly by applying the NOILC scheme, with a reduced root-mean-square (RMS) error from $\unit[13]{^\circ}$ to less than $\unit[2]{^\circ}$ after 30 iterations as can be seen in Fig. \ref{fig:error}. The non-causality of the NOILC scheme allows the elimination of the delay between the angle response and the set point, which occurs when no learning control is used. 

When the feedback controller is used without any learning, the robotic arm bounces back slightly after an initial steep response when commanding an angle jump of $\unit[60]{^\circ}$. The reason for this is that it compresses the deflating actuator, causing the pressure in this actuator to temporarily increase and hence the arm to bounce back. Note that if proportional valves were deployed, the temporary pressure increase in the compressed actuator would be less pronounced because of their higher flow capacity. Considering the full pressure dynamics on the other hand (i.e. not simplifying \eqref{eq:pressDyn}) would not avoid the pressure build-up, because the valves of the deflating actuator are already fully open. The NOILC scheme compensates for this effect by increasing the pressure difference for a short period, which can be seen by the peaks in the total pressure difference applied in iteration 30 (red curve) occurring at $t=\{1.6, 2.6, 3.6, \unit[4.6]{s}\}$. 

Note that if the same control input is applied, the disturbances are repeatable over different iterations of the learning scheme since they arise from unmodeled dynamics. Over a single iteration, the behavior is also repeatable, as can be seen for example by the similar angular response when commanding a set point jump from $\unit[-30]{^{\circ}}$ to $\unit[30]{^{\circ}}$ occurring at $t=\{2.6, \unit[4.6]{s}\}$.

A comparison between the commanded pressure differences reveals that the feedback controller without learning has to rely on integral action to reach the set point, while the NOILC directly adjusts the pressure difference required for a certain angular set point.

\vspace{-2pt}
\section{Conclusion}\label{sec:Conclusion}
A norm-optimal ILC scheme has been presented to improve the position tracking performance with an articulated soft robotic arm. The non-causal learning scheme is based on a simple model of the dynamics, which guides the learning and brings advantages compared to simpler ILC approaches, e.g. PD-type ILC. As opposed to a model-based feedback controller, the tracking performance of the learning scheme is not limited by the model accuracy. Results have shown that performance can be improved significantly for an aggressive trajectory. Learning a correction signal for different reference trajectories is a practical alternative to both more expensive valves and more advanced causal control approaches. Future work includes the design of a fully inflatable system and will investigate how learning control generalizes to systems with multiple degrees of freedom.
\vspace{-4pt}
\section*{Acknowledgment}
The authors would like to thank Michael Egli, Daniel Wagner and Marc-Andr\`{e} Corzillius for their contribution to the development of the prototype. 
\vspace{-4pt}

\bibliographystyle{IEEEtran}
\bibliography{bibliography}

\end{document}